\pdfoutput=1

\documentclass[11pt,a4paper]{article}
\usepackage[hyperref]{acl2020}
\usepackage{times}
\usepackage{latexsym}
\usepackage{booktabs}
\usepackage{tikz-dependency}
\usepackage{adjustbox}
\usepackage{multirow}
\usepackage{amsmath}

\usepackage{microtype}

\newcommand{\emldisplay}[2]{\texttt{\href{mailto:#1}{#2}}}

\aclfinalcopy % Uncomment this line for the final submission
\def\aclpaperid{27} %  Enter the acl Paper ID here

\title{K{\o}psala: Transition-Based Graph Parsing
via \\ Efficient Training and Effective Encoding}

\author{Daniel Hershcovich\thanks{\hspace{2pt} Equal contribution}$\;^{\diamondsuit}$
   \quad
  Miryam de Lhoneux$^{*\diamondsuit}$ \quad
  Artur Kulmizev$^\heartsuit$ \\ \bfseries
  Elham Pejhan$^\diamondsuit$ \quad
  Joakim Nivre$^\heartsuit$ \\
  $^\diamondsuit$University of Copenhagen \quad $^\heartsuit$Uppsala University \\
  \{\emldisplay{dh@di.ku.dk}{dh},
  \emldisplay{ml@di.ku.dk}{ml}, 
  \emldisplay{ep@di.ku.dk}{ep}\}\texttt{@di.ku.dk}, \\
  \{\emldisplay{artur.kulmizev@lingfil.uu.se}{artur.kulmizev}, \emldisplay{joakim.nivre@lingfil.uu.se}{joakim.nivre}\}\texttt{@lingfil.uu.se}
}

\date{}

\begin{document}
\maketitle
\begin{abstract}
We present K{\o}psala, the Copenhagen-Uppsala system for the Enhanced Universal Dependencies Shared Task at IWPT 2020. Our system is a pipeline consisting of off-the-shelf models for everything but enhanced graph parsing, and for the latter, a transition-based graph parser adapted from \citet{che-etal-2019-hit}. We train a single enhanced parser model per language, using gold sentence splitting and tokenization for training, and rely only on tokenized surface forms and multilingual BERT for encoding. While a bug introduced just before submission resulted in a severe drop in precision, its post-submission fix would bring us to 4th place in the official ranking, according to average ELAS. Our parser demonstrates that a unified pipeline is effective for both Meaning Representation Parsing and Enhanced Universal Dependencies.
\end{abstract}

\section{Introduction}
\label{sec:introduction}

\begin{figure}[t]
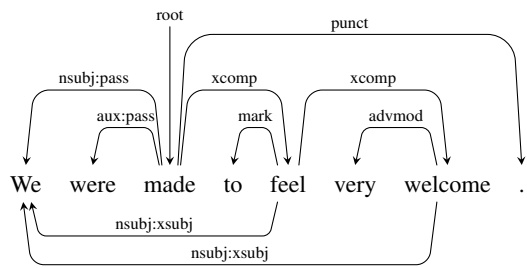

  \centering
    \begin{dependency}[text only label, label style={above}, font=\small]
    \begin{deptext}[column sep=.5em,ampersand replacement=\^]
    We \^ were \^ made \^ to \^ feel \^ very \^ welcome \^ . \\
    \end{deptext}
        \depedge[edge start x offset=1pt]{3}{1}{nsubj:pass}
        \depedge[edge below,edge unit distance=.65ex,edge end x offset=2pt]{5}{1}{nsubj:xsubj}
        \depedge[edge below,edge unit distance=.8ex,edge end x offset=-1pt]{7}{1}{nsubj:xsubj}
        \depedge{3}{2}{aux:pass}
        \deproot[edge unit distance=3ex]{3}{root}
        \depedge{5}{4}{mark}
        \depedge{3}{5}{xcomp}
        \depedge{7}{6}{advmod}
        \depedge{5}{7}{xcomp}
        \depedge[edge unit distance=2ex,edge start x offset=-1pt]{3}{8}{punct}
    \end{dependency}
\caption{ED for \texttt{reviews-077034-0002} from \texttt{UD\_English-EWT},
containing a control verb (made).
Arcs above the sentence are also in BD.\label{fig:example_english}}
\end{figure}

\begin{figure*}[ht]
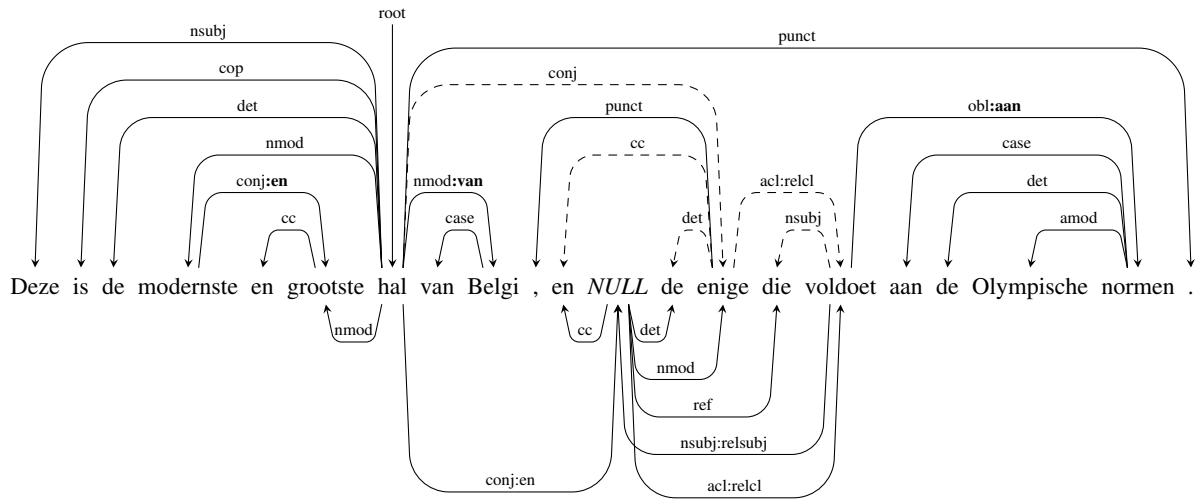

  \centering
\begin{dependency}[text only label, label style={above}, font=\small]
\begin{deptext}[column sep=0em]
Deze \& is  \& de  \& modernste \& en    \& grootste \& hal  \& van \& België \& ,     \& en  \& \textit{NULL}  \& de  \& enige \& die  \& voldoet \& aan \& de  \& Olympische \& normen \& .     \\
\end{deptext}
\depedge{7}{1}{nsubj}
\depedge{7}{2}{cop}
\depedge{7}{3}{det}
\depedge{7}{4}{nmod}
\depedge{6}{5}{cc}
\depedge{4}{6}{conj\textbf{:en}}
\depedge[edge below]{7}{6}{nmod}
\deproot[edge unit distance=5ex]{7}{root}
\depedge{9}{8}{case}
\depedge{7}{9}{nmod\textbf{:van}}
\depedge{14}{10}{punct}
\depedge[dashed]{14}{11}{cc}
\depedge[edge below]{12}{11}{cc}
\depedge[edge below]{7}{12}{conj:en}
\depedge[edge below]{16}{12}{nsubj:relsubj}
\depedge[dashed]{14}{13}{det}
\depedge[edge below]{12}{13}{det}
\depedge[dashed,edge unit distance=2ex]{7}{14}{conj}
\depedge[edge below]{12}{14}{nmod}
\depedge[dashed]{16}{15}{nsubj}
\depedge[edge below]{12}{15}{ref}
\depedge[dashed]{14}{16}{acl:relcl}
\depedge[edge below,edge unit distance=3.7ex]{12}{16}{acl:relcl}
\depedge{20}{17}{case}
\depedge{20}{18}{det}
\depedge{20}{19}{amod}
\depedge{16}{20}{obl\textbf{:aan}}
\depedge[edge unit distance=1.2ex]{7}{21}{punct}
\end{dependency}
\caption{\texttt{\small wiki-3745.p.38.s.5} from \texttt{UD\_Dutch-LassySmall}, containing a null node \textit{NULL}, not in the original sentence,
coordination and case suffixes (\texttt{\textbf{:en}}, \texttt{\textbf{:van}}, \texttt{\textbf{:aan}}),
and propagation of conjuncts (hal~$\to$~grootste).
The dashed edges are deleted in ED, and the edges below the sentence are added.
Note the cycle \textit{NULL}~$\leftrightarrow$~voldoet.
}
\label{fig:example_dutch}
\end{figure*}

The IWPT 2020 Shared Task on Parsing into Enhanced Universal Dependencies
\cite{EUDparsingST:2020}
involves sentence segmentation, tokenization, lemmatization,
part-of-speech tagging,
morphological analysis, basic dependency parsing, and finally
(for the first time)
\textit{enhanced} dependency parsing.
The enhancements encode
case information, elided predicates,
and shared arguments due to conjunction, control, raising
and relative clauses
(see Figures~\ref{fig:example_english} and~\ref{fig:example_dutch}).

In Universal Dependencies v2 \cite[UD;][]{nivre2020universal},
enhanced dependencies (ED) are a separate
dependency graph than the \textit{basic} dependency tree (BD).
However, ED is \textit{almost} a super-set of BD,\footnote{Some BD arcs are deleted in ED, e.g., \texttt{orphan} arcs.}
and so
most previous approaches \cite{SCHUSTER16.779,nivre-etal-2018-enhancing} have attempted to recover ED from BD using language-specific rules.
On the other hand,
\citet{hershcovich-etal-2018-universal}
experimented with TUPA, a transition-based directed acyclic graph (DAG) parser originally designed for parsing UCCA \cite{abend-rappoport-2013-universal}, for supervised ED parsing. They converted ED to UCCA-like graphs and did not use pre-trained contextualized embeddings,
yielding sub-optimal results.
Taking a similar approach, we adapt a transition-based graph parser
\cite{che-etal-2019-hit} designed for Meaning Representation Parsing \cite{oepen-etal-2019-mrp}, but parse ED directly
and use BERT embeddings \cite{devlin-etal-2019-bert}.

The main contribution of our work is a transition system
supporting the graph structures exhibited by ED,
including null nodes (meaning this is not a strictly
bilexical formalism), cycles and non-crossing graphs
(\S\ref{sec:transition_system}),
as Figure~\ref{fig:transition_sequence} demonstrates
for the sentence from Figure~\ref{fig:example_dutch}.
We parse ED completely separately from BD,
demonstrating the applicability of a full graph parser, starting
from only segmented and tokenized text, to ED.
Our code is available at \url{https://github.com/coastalcph/koepsala-parser}.

\section{Preprocessing}
\label{sec:preprocessing}
As the focus of this shared task is ED parsing, we rely on existing systems for preprocessing. Here, we consider two off-the-shelf pipelines: \textsc{stanza} \cite{qi2020stanza}\footnote{\url{https://stanfordnlp.github.io/stanza/models.html}} and \textsc{udpipe} 1.2 \cite{udpipe:2017,straka2016udpipe},\footnote{\url{https://lindat.mff.cuni.cz/repository/xmlui/handle/11234/1-3131}} both of which have models pre-trained on UD v2.5 treebanks.
We experiment with either pipeline during prediction to process the raw text files for the dev and test sets,
eventually selecting \textsc{udpipe} for our primary
submission.
This process entails sentence segmentation, tokenization, lemmatization, part-of-speech tagging, morphological feature tagging, and BD parsing.\footnote{The preprocessing output, except for segmentation and tokenization,
is not used in any way by the ED parser, since it just
uses BERT for token representation (\S\ref{sec:classifier}).}
For training our ED parser (\S\ref{sec:parser}), however,
we use gold inputs for simplicity.
We use the \texttt{conllu} Python package\footnote{\url{https://github.com/EmilStenstrom/conllu}} to read \mbox{CoNLL-U} files.

\paragraph{Preprocessing model selection.}
Since the dev and test data do not denote their source treebanks, we simply process the text using the pipeline model trained on the language's largest treebank. 
To experiment with an alternative method, for languages with more than one treebank, we also train \textsc{udpipe} models on combined training treebanks. Table \ref{table:trained_udpipe_results} shows the comparison of LAS on the combined dev set, for these models and for the models (pre-)trained on the language's largest treebank.
The results show that using the combined training sets does not lead to consistent improvements in terms of LAS, and so we continue using pre-trained treebank-specific preprocessing models henceforth.

\begin{table}[t]
\centering
\begin{tabular}{@{}rrccccc@{}}
\toprule
                & \multicolumn{4}{c}{\textbf{Language}} \\
  & \small\textsc{Czech} & \small\textsc{Dutch} & \small\textsc{Estonian} & \small\textsc{Polish} \\ \midrule
{combined}      & 78.88 & \textbf{76.50} & 77.01 &  \textbf{82.96} \\
{largest}     & \textbf{83.97} & 74.97 & \textbf{77.61} &  82.59 \\
\bottomrule
\end{tabular}
\caption{LAS on the combined dev set for \textsc{udpipe} models trained on the language's combined training treebanks and the models trained on the language's largest treebank.
No consistent trend is observed.}
\label{table:trained_udpipe_results}
\end{table}

\section{Transition-Based Enhanced Dependency Parser}
\label{sec:parser}

Our system is a transition-based graph parser,
based on the HIT-SCIR system \cite{che-etal-2019-hit},
which achieved the highest average score across frameworks (AMR, EDS, UCCA, DM and PSD) in the
CoNLL 2019 shared task on Meaning Representation Parsing
\cite[MRP;][]{oepen-etal-2019-mrp}.
It is written in the AllenNLP \cite{Gardner2017AllenNLP}
framework.
For training efficiently, it employs stack LSTMs \cite{dyer-etal-2015-transition}, batching operations across sentences.
For better encoding, HIT-SCIR fine-tuned BERT \cite{devlin-etal-2019-bert} while training the parser.

A transition-based parser operates by manipulating a buffer (originally containing the input words provided by the preprocessor, see~\S\ref{sec:preprocessing}) and a stack (originally containing the root, i.e., word at index 0), to incrementally create the output dependency graph.
At each point in the parsing process, a transition is selected out of a pre-defined set of possible transitions.
A classifier is trained to predict the best transition to apply at each step, by mimicking an oracle during training (see \S\ref{sec:transition_system}).

HIT-SCIR used a different transition system per framework
(AMR, EDS, UCCA; and one system for DM and PSD),
according to the graph properties of each
and based on existing framework-specific parsers
\cite{liu-etal-2018-amr,buys-blunsom-2017-robust,hershcovich-etal-2017-transition,wang-etal-2018-neural}.
We construct a transition system for ED using subsets of transitions from two of the HIT-SCIR systems: their system for DM and PSD, as well as their system for UCCA, with some further adaptations specific to ED graphs.

\subsection{Transition System}\label{sec:transition_system}

Our system contains the following transitions: \{\textsc{Shift}, \textsc{Left-Edge$_l$}, \textsc{Right-Edge$_l$}, \textsc{Reduce-0}, \textsc{Reduce-1}, \textsc{Node}, \textsc{Swap} and \textsc{Finish}\}.
The \textsc{Shift} transition pops the first element of the buffer and pushes it onto the stack. The \textsc{Left-Edge$_l$} and \textsc{Right-Edge$_l$} transitions add an arc\footnote{For consistency, we keep the transition nomenclature using ``\textsc{Edge}'', although they create directed dependency \textit{arcs}.
Note that in analogous transitions in some transition systems, such as \textsc{ArcEager} \cite{nivre-2003-efficient}, the dependent of the transition is also popped out of the stack as part of either of these two transitions. Here, since dependents can have multiple heads and can have arcs with multiple labels, we stick to the \textsc{Edge} action and use our two \textsc{Reduce} transitions to pop elements of the stack when necessary.}
between the two top items of the stack with label $l$.
We need two different \textsc{Reduce} transitions to pop the topmost and second topmost items of the stack, which we name \textsc{Reduce-0} and a \textsc{Reduce-1} respectively. This makes it possible to construct length-2 cycles, which ED allows (and most MRP frameworks do not). The \textsc{Node} transition inserts a null node as the first element of the buffer, needed to support null nodes.
\textsc{Swap} moves the second-top node of the stack to the buffer, thus swapping the order between the two top nodes of the stack. This is necessary for handling crossing graphs (analogous to non-projective trees). Finally, \textsc{Finish} terminates the transition sequence.
A formal definition of the transition set is shown in Figure~\ref{fig:transitions}.

\begin{figure*}[t]
	\begin{adjustbox}{width=\textwidth}
	\begin{tabular}{llll|l|llllc|c}
	\toprule
		\multicolumn{4}{c|}{\textbf{\small Before Transition}} & \textbf{\small Transition} & \multicolumn{5}{c|}{\textbf{\small After Transition}} & \textbf{\small Condition} \\
		\textbf{\footnotesize Stack} & \textbf{\footnotesize Buffer} & \textbf{\footnotesize Nodes} & \textbf{\footnotesize Arcs} & & \textbf{\footnotesize Stack} & \textbf{\footnotesize Buffer} & \textbf{\footnotesize Nodes} & \textbf{\footnotesize Arcs} & \textbf{\footnotesize Terminal?} & \\\midrule
		$\Sigma$ & $b \;|\; B$ & $V$ & $A$ & \textsc{Shift} & $\Sigma \;|\; b$ & $B$ & $V$ & $A$ & $-$ & \\
        $\Sigma \;|\; s_0$ & $B$ & $V$ & $A$ & \textsc{Reduce-0} & $\Sigma$ & $B$ & $V$ & $A$ & $-$ & $s_0 \neq \mathrm{root} \wedge (\cdot,s_0)_\cdot\in A$\\
        $\Sigma \;|\; s_1, s_0 $ & $B$ & $V$ & $A$ & \textsc{Reduce-1} & $\Sigma\;|\; s_0$ & $B$ & $V$ & $A$ & $-$ & $s_1 \neq \mathrm{root} \wedge (\cdot,s_1)_\cdot\in A$\\
		$\Sigma$ & $B$ & $V$ & $A$ & \textsc{Node} & $\Sigma$ & $b \;|\; B$ & $V \cup \{ b \}$ & $A$ & $-$ &
		\\
        $\Sigma \;|\; s_1,s_0$ & $B$ & $V$ & $A$ & \textsc{Left-Edge$_l$} & $\Sigma \;|\; s_1,s_0$ & $B$ & $V$ & $A \cup \{ (s_0,s_1)_l \}$ & $-$ & $s_1 \neq \mathrm{root} \wedge (s_0,s_1)_l \not\in A$\\
		$\Sigma \;|\; s_1,s_0$ & $B$ & $V$ & $A$ & \textsc{Right-Edge$_l$} & $\Sigma \;|\; s_1,s_0$ & $B$ & $V$ & $A \cup \{ (s_1,s_0)_l \}$ & $-$ &  $(s_1,s_0)_l \not\in A$ \\
        $\Sigma \;|\; s_1,s_0$ & $B$ & $V$ & $A$ & \textsc{Swap} & $\Sigma \;|\; s_0$ & $s_1 \;|\; B$ & $V$ & $A$ & $-$ & 
        $s_1 \neq \mathrm{root} \wedge \mathrm{i}(s_1) < \mathrm{i}(s_0)$ \\
		$[\mathrm{root}]$ & [ ] & $V$ & $A$ & \textsc{Finish} & [ ] & [ ] & $V$ & $A$ & $+$ & \\\bottomrule
	\end{tabular}
	\end{adjustbox}
	\caption{\label{fig:transitions}
	  Our transition set.
	  We write the stack with its top to the right and the buffer with its head to the left.
	  $(h,d)_l$ denotes an $l$-labeled dependency with head $h$ and dependent $d$. 
	  $\mathrm{i}(x)$ is the generated order (see \S\ref{sec:transition_system}).
	}
\end{figure*}

Separate \textsc{Edge} transitions exist for each edge label.
Labels containing coordination or case suffixes
(such as \texttt{nmod:van}) are
treated as any other label and are not split,
resulting in a large number of transitions for some
languages, shown in Table~\ref{tab:transition_counts}.

\textsc{Node} transitions, on the other hand, do not select
any label or features, since null nodes are only evaluated
with respect to their incoming and outgoing edges.
All other information is ignored, and thus not predicted by
the parser: predicted null nodes are thus only placeholders.

\begin{table}[t]
    \centering
    \begin{tabular}{l|rrr}
\toprule
\textbf{Language}   & Total & \textsc{Edge} & w/ suffix \\ \midrule
\small\textsc{Arabic}     &402 &395&345\\
\small\textsc{Bulgarian}  &197 &191&137\\
\small\textsc{Czech}      &768 &761&702\\
\small\textsc{Dutch}      &393 &386&336\\
\small\textsc{English}    &300 &293&232\\
\small\textsc{Estonian}   &445 &438&381\\
\small\textsc{Finnish}    &266 &259&210\\
\small\textsc{French}     &112 &106&59 \\
\small\textsc{Italian}    &281 &274&216\\
\small\textsc{Latvian}    &238 &232&161\\
\small\textsc{Lithuanian} &323 &317&263\\
\small\textsc{Polish}     &676 &669&615\\
\small\textsc{Russian}    &944 &938&861\\
\small\textsc{Slovak}     &266 &259&204\\
\small\textsc{Swedish}    &209 &202&153\\
\small\textsc{Tamil}      &146 &140&103\\
\small\textsc{Ukrainian}  &290 &283&225\\ \bottomrule
\end{tabular}
    \caption{Number of transitions for each language.}
    \label{tab:transition_counts}
\end{table}

\paragraph{Constraints.}

In addition to the modified transition set, we change the constraints for some transitions according to the required graph structure.

Since \textsc{Left-Edge$_l$} and \textsc{Right-Edge$_l$} transitions do not reduce the dependent, we need to ensure that we do not draw the same arc twice. For this reason, these transitions are not allowed if there is already an arc with label $l$ between the two nodes. We also disallow to add an arc with the root as dependent.

To ensure every node gets attached to at least one head, we disallow the \textsc{Reduce-0} and \textsc{Reduce-1} transitions for nodes that do not have a head yet. 
We also disallow reducing the root.

For the \textsc{Swap} transition, we maintain the \textit{generated order} of each node, assigned when the node is shifted (for words) or created (for null nodes).
To prevent infinite loops during inference, we only allow swapping nodes whose order in the stack is the same as their generation order.

To limit repeated actions, we arbitrarily constrain \textsc{Node} transitions such that there are no more null nodes than words
(although a lower limit would suffice), and
\textsc{Edge} transitions to limit the number of heads per node to 7.\footnote{While the observed number of heads per node in the data goes up to 36, in the training data there is only a small minority of cases where a node has more than 7 heads.}

\textsc{Finish} is only allowed when the buffer is empty and the stack only contains the root.
If no valid transition is available, the sequence is
terminated prematurely by applying the \textsc{Finish} transition, regardless of the \textsc{Finish} constraints.

\paragraph{Oracle.}
We use a static oracle similar to HIT-SCIR
(a single ``gold'' transition sequence is given during training, which the parser is forced to follow), but develop one for our transition system.

The oracle deterministically chooses the transition to take given the current configuration. Let $s_1$ and $s_0$ be the two top items of the stack and $b$ the first item of the buffer (if these are defined in the current configuration). If the buffer is empty and the stack only contains the root, take a \textsc{Finish} transition. Otherwise, if there is an arc between $s_1$ and $s_0$ with label $l$ that has not yet been constructed, take the necessary \textsc{Right-Edge$_l$} or \textsc{Left-Edge$_l$} action. Otherwise, if $s_0$ has a node dependent, take a \textsc{Node} transition. Otherwise, if $s_0$ has all its heads and dependents, take \textsc{Reduce-0}, if $s_1$ has all its heads and dependents, take \textsc{Reduce-1}. Otherwise, if $s_1$ and $s_0$ are in their generated order and $s_0$ has a head or a dependent in the stack that is not $s_1$, take a \textsc{Swap}. Otherwise \textsc{Shift}.
Figure~\ref{fig:transition_sequence} shows an example transition sequence.

\begin{figure*}[t]
  \small\centering
  \setlength{\tabcolsep}{8pt}
  \begin{tabular}{@{}lllll@{}}
  \toprule
    & \textbf{Transition} & \textbf{Stack} & \textbf{Buffer} & \textbf{Arc added} \\\midrule
     &  & $[$ ROOT $]$ & $[$ Deze is (\dots)$]$ & \\
    1-6 & \textsc{Shift} & $[$ (\dots) en grootste $]$ & $[$ hal van (\dots) $]$ & \\
    7 & \textsc{Left-Edge}$_\texttt{cc}$ & $[$ (\dots) en grootste $]$ & $[$ hal van (\dots) $]$ & $($grootste, en$)_\texttt{cc}$\\
    8 & \textsc{Reduce-1} & $[$ (\dots) modernste grootste $]$ & $[$ hal van (\dots) $]$ & \\
    9 & \textsc{Right-Edge}$_\texttt{conj:en}$ & $[$ (\dots) modernste grootste $]$ & $[$ hal van (\dots) $]$ & $($grootste, modernste$)_\texttt{conj:en}$\\
    10 & \textsc{Shift} & $[$ (\dots) grootste hal $]$ & $[$ van België (\dots) $]$ & \\
    11 & \textsc{Left-Edge}$_\texttt{nmod}$ & $[$ (\dots) grootste hal $]$ & $[$ van België (\dots) $]$ & $($hal, grootste$)_\texttt{nmod}$\\
    12 & \textsc{Node} & $[$ (\dots) grootste hal $]$ & $[$ \textit{NULL} van (\dots) $]$ & \\\midrule
    13-21 & \multicolumn{4}{c}{Series of \textsc{Left-Edge} and \textsc{Reduce-1} transitions}\\\midrule
    22 & \textsc{Right-Edge}$_\texttt{root}$ & $[$ ROOT hal $]$ & $[$ \textit{NULL} van (\dots) $]$ & $($ROOT, hal$)_\texttt{root}$\\
    23 & \textsc{Shift} & $[$ ROOT hal \textit{NULL} $]$ & $[$ van België (\dots) $]$ & \\
    24 & \textsc{Right-Edge}$_\texttt{conj:en}$ & $[$ ROOT hal \textit{NULL} $]$ & $[$ van België (\dots) $]$ & $($hal, \textit{NULL}$)_\texttt{conj:en}$\\
    25-26 & \textsc{Shift} & $[$ (\dots) van België $]$ & $[$ , en (\dots) $]$ & \\
    27 & \textsc{Left-Edge}$_\texttt{case}$ & $[$  (\dots) van België $]$ & $[$ , en (\dots) $]$ & $($België, van$)_\texttt{case}$\\
    28 & \textsc{Reduce-1} & $[$  (\dots) \textit{NULL} België $]$ & $[$ , en (\dots) $]$ & \\
    29 & \textsc{Swap} & $[$ ROOT hal België $]$ & $[$ \textit{NULL} , (\dots) $]$ & \\
    30 & \textsc{Right-Edge}$_\texttt{nmod:van}$ & $[$  (\dots) hal België $]$ & $[$ \textit{NULL} , (\dots) $]$ & $($hal, België$)_\texttt{nmod:van}$\\
    31 & \textsc{Reduce-0} & $[$  ROOT hal $]$ & $[$ \textit{NULL} , (\dots) $]$ & \\
    32-34 & \textsc{Shift} & $[$  (\ldots) , en $]$ & $[$ de enige (\dots) $]$ & \\
    35 & \textsc{Swap} & $[$  (\ldots) \textit{NULL} en $]$ & $[$ , de (\dots) $]$ & \\
    36 & \textsc{Right-Edge}$_\texttt{cc}$ & $[$  (\ldots) \textit{NULL} en $]$ & $[$ , de (\dots) $]$ & $($\textit{NULL}, en$)_\texttt{cc}$\\
    37 & \textsc{Reduce-0} & $[$  ROOT hal \textit{NULL} $]$ & $[$ , de (\dots) $]$ & \\
    38-39 & \textsc{Shift} & $[$  (\dots) , de $]$ & $[$ enige die (\dots) $]$ & \\
    40 & \textsc{Swap} & $[$  (\dots) \textit{NULL} de $]$ & $[$ , enige (\dots) $]$ & \\
    41 & \textsc{Right-Edge}$_\texttt{det}$ & $[$  (\dots) \textit{NULL} de $]$ & $[$ , enige (\dots) $]$ & $($\textit{NULL}, de$)_\texttt{det}$\\
    42 & \textsc{Reduce-0} & $[$  ROOT hal \textit{NULL} $]$ & $[$ , enige (\dots) $]$ & \\
    43-44 & \textsc{Shift} & $[$  (\dots) , enige $]$ & $[$ die voldoet (\dots) $]$ & \\
    45 & \textsc{Left-Edge}$_\texttt{punct}$ & $[$  (\dots) , enige $]$ & $[$ die voldoet (\dots) $]$ & $($enige, ,$)_\texttt{punct}$\\
    46 & \textsc{Reduce-1} & $[$  (\dots) \textit{NULL} enige $]$ & $[$ die voldoet (\dots) $]$ & \\
    47 & \textsc{Right-Edge}$_\texttt{nmod}$ & $[$  (\dots) \textit{NULL} enige $]$ & $[$ die voldoet (\dots) $]$ & $($\textit{NULL}, enige$)_\texttt{nmod}$\\
    48 & \textsc{Reduce-0} & $[$  ROOT hal \textit{NULL} $]$ & $[$ die voldoet (\dots) $]$ & \\
    49 & \textsc{Shift} & $[$  (\dots) \textit{NULL} die $]$ & $[$ voldoet aan (\dots) $]$ & \\
    50 & \textsc{Right-Edge}$_\texttt{ref}$ & $[$  (\dots) \textit{NULL} die $]$ & $[$ voldoet aan (\dots) $]$ & $($\textit{NULL}, die$)_\texttt{ref}$\\
    51 & \textsc{Reduce-0} & $[$  ROOT hal \textit{NULL} $]$ & $[$ voldoet aan (\dots) $]$ & \\
    52 & \textsc{Shift} & $[$  (\dots) \textit{NULL} voldoet $]$ & $[$ aan de (\dots) $]$ & \\
    53 & \textsc{Right-Edge}$_\texttt{acl:relcl}$ & $[$  (\dots) \textit{NULL} voldoet $]$ & $[$ aan de (\dots) $]$ & $($\textit{NULL}, voldoet$)_\texttt{acl:relcl}$\\
    54 & \textsc{Left-Edge}$_\texttt{nsubj:relsubj}$ & $[$  (\dots) \textit{NULL} voldoet $]$ & $[$ aan de (\dots) $]$ & $($voldoet, \textit{NULL}$)_\texttt{nsubj:relsubj}$\\\midrule
    55-69 & \multicolumn{4}{c}{(\dots)}\\\midrule
    70 & \textsc{Right-Edge}$_\texttt{punct}$ & $[$  ROOT hal . $]$ & $[$ $]$ & $($hal, .$)_\texttt{punct}$\\
    71-72 & \textsc{Reduce-0} & $[$  ROOT $]$ & $[$ $]$ & \\
    73 & \textsc{Finish} & $[$  ROOT $]$ & $[$ $]$ & \\\bottomrule
  \end{tabular}
  \caption{Oracle transition sequence for the sentence from Figure~\ref{fig:example_dutch}. Consecutive \textsc{Shift}s grouped for brevity.}
  \label{fig:transition_sequence}
\end{figure*}

\subsection{Classifier}\label{sec:classifier}

The parser uses BERT \cite{devlin-etal-2019-bert}
for token representation. While \citet{che-etal-2019-hit} used pre-trained English model (\texttt{wwm\_cased\_L-24\_H-1024\-16}),
we replaced it with a pre-trained multilingual one (\texttt{multi\_cased\_L-12\_H-768\-12}),\footnote{\url{https://github.com/google-research/bert/blob/master/multilingual.md}}
trained on 104 languages, including all 17 languages participating
in the shared task.
As done by \citet{che-etal-2019-hit}, we use the
\texttt{bert-pretrained} text field embedder from AllenNLP,
which extracts the first word-piece of each token, applying a
scalar mix on all layers of transformer.

The transition classifier is a stack-LSTM \cite{dyer-etal-2015-transition} with only BERT embedding features for words,
as well as a scalar feature denoting the ratio
between the number of (null) nodes and the number of words
\cite{hershcovich-etal-2017-transition},
as in HIT-SCIR.
We do not fine-tune BERT due to memory limitations,
though fine-tuning would likely result in improved performance.

\subsection{Postprocessing}\label{sec:postprocessing}

The enhanced graphs are required to be connected,
i.e., every node must be reachable from the root.\footnote{This is enforced by the task organizers by running \texttt{validate.py --level 2 --lang ud} on the system predictions before evaluation.}
While the transition constraints ensure that every node has a head,
there may be unconnected \textit{cycles} at the end of
the parse, resulting in invalid graphs.
To fix the problem, at the end of the parse, we iteratively
find the unconnected node with the most descendants,
and attach it to the predicate (the first dependent of the root) with an \texttt{orphan}-labeled
arc.
In addition to unconnected cycles, this resolves the problem
of prematurely terminated transition sequences due to
no valid transition being available according to the constraints:
instead of resulting in partially-constructed graphs,
headless nodes are similarly attached with an \texttt{orphan}-labeled
arc to the predicate, if it exists, or otherwise to the root.

\paragraph{Parsing tragedy.}
Our postprocessing procedure to attach unconnected subgraphs
had a bug at the time of submission, where many nodes were incorrectly identified as unconnected and thus unnecessarily attached to the predicate/root.
While this still yielded valid graphs,
precision dropped precipitously from before the introduction
of the postprocessing procedure.
Due to the late stage in the evaluation period at which
we made this change, we failed to properly monitor our
development scores and could not identify the cause for
the drop in time, resulting in low official scores.
However, after submission we identified the bug and fixed it,\footnote{\url{https://github.com/coastalcph/koepsala-parser/commit/1b872ad9fc2652649c11eb0a8622c744c92e8cbb}}
improving our parser's accuracy back to the range we had
observed during development.

\subsection{Training}\label{sec:training}

For training the ED parser
we do \textit{not} simply train it on the largest treebank per language, but rather train it on the concatenated training treebanks per language.
In preliminary experiments, this did lead to improvements in terms of combined dev ELAS over treebank-specific models,
contrary to our findings in BD parsing for preprocessing (\S\ref{sec:preprocessing}).
We train our models on an NVIDIA P100 GPU with a batch size of 8. All other hyperparameters can be found in the configuration files in the repository.\footnote{\url{https://github.com/coastalcph/koepsala-parser/blob/master/config/transition_eud.jsonnet}}

Training until convergence took 1h30 (for Tamil, the smallest treebank) to up to 2 days (for Arabic, which contains many long sentences).
Prediction on the dev set took between 4 minutes (for Tamil)
and 55 minutes (for Czech),
ranging from 117 words/second (7 sentences/second, for Tamil) to
1300 words/second (81 sentences/second, for Czech),
including the model loading time.

\subsection{Baselines}\label{sec:baseline}

In addition to providing validation scores for our trained parsers, we consider three competitive baselines, as provided by the task organizers:

\begin{itemize}
    \item \texttt{B1}: gold standard dependency trees copied as enhanced graphs. Though this can be technically considered an upper bound, as gold tree information is provided, it should nonetheless provide some idea of how much of the enhanced graph can be derived from the dependency tree. 
    \item \texttt{B2}: predicted trees yielded by UDPipe models trained on UD v2.5 (using the largest treebank where applicable), copied as enhanced graphs. This is more representative than \texttt{B1} of realistic parsing scenarios, which rely on predictions. 
    \item \texttt{B3}: similar to \texttt{B2}, but applying the Stanford enhancer post-hoc over the predicted trees. Scores for Finnish and Latvian were not provided. 
\end{itemize}

\section{Results}
\label{sec:results}
\begin{table}[t]
\centering
\setlength{\tabcolsep}{4pt}
\begin{tabular}{@{}rccccc@{}}
\toprule
        \textbf{Language}   & \multicolumn{3}{c}{\textbf{Baselines}} & \multicolumn{2}{c}{\textbf{Ours}} \\ \midrule
               & \textit{B1}       & \textit{B2}       & \textit{B3}      & \textit{official}    & \textit{fixed}    \\ \midrule
\small\textsc{Arabic}     & 67.35    & 46.41    & 45.16   & 60.84         &   69.51       \\
\small\textsc{Bulgarian}  & 85.82    & 73.74    & 79.9    & 68.88         &     84.49     \\
\small\textsc{Czech}      & 78.44    & 65.31    & 63.62   & 61.11         &     74.79     \\
\small\textsc{Dutch}      & 82.48    & 62.97    & 72.65   & 62.93         &     76.92     \\
\small\textsc{English}    & 84.30    & 66.83    & 76.16   & 65.37         &      81.05    \\
\small\textsc{Estonian}   & 76.38    & 57.53    & 54.34   & 59.07         &       72.38   \\
\small\textsc{Finnish}    & 78.26    & 61.71    & -       & 67.54         &       81.58   \\
\small\textsc{French}     & 97.49    & 71.14    & 63.31   & 67.93         &       82.76   \\
\small\textsc{Italian}    & 80.20    & 70.33    & 83.03   & 69.08         &       84.66   \\
\small\textsc{Latvian}    & 79.31    & 59.14    & -       & 64.75         &       79.12   \\
\small\textsc{Lithuanian} & 74.22    & 46.78    & 44.84   & 56.28         &        69.09  \\
\small\textsc{Polish}     & 81.59    & 66.38    & 65.37   & 61.34         &      73.89    \\
\small\textsc{Russian}    & 79.63    & 68.33    & 67.80   & 64.23         &      78.90    \\
\small\textsc{Slovak}     & 77.60    & 60.02    & 58.05   & 64.08         &      77.44    \\
\small\textsc{Swedish}    & 80.98    & 62.18    & 71.53   & 64.50         &       78.61   \\
\small\textsc{Tamil}      & 76.29    & 40.71    & 40.25   & 47.44         &      56.85    \\
\small\textsc{Ukrainian}  & 77.24    & 58.73    & 56.92   & 64.17         &      78.10    \\ \midrule
\small\textsc{\textbf{Average}}    & 79.86    & 61.07    & 62.90   & 62.91         &    76.48      \\ \bottomrule
\end{tabular}
\caption{Main results for Enhanced Universal Dependencies shared task (ELAS), as evaluated on the provided test sets. \textit{B1, B2, B3} refer to organizer-provided baseline systems. \textit{official} refers to our official submission, prior to fixing the unconnected graph issue (\textit{fixed}). }
\label{table:main_results}
\end{table}

Table \ref{table:main_results} displays our results on the per-language (not per-treebank) test partitions of the shared task data.
As explained in \S\ref{sec:preprocessing}, for languages with multiple training treebanks available (Czech, Estonian, Dutch, Polish), we preprocessed the raw text of each treebank using the pipeline trained on the \textit{largest} treebank available for that language (e.g. \texttt{alpino} for Dutch).
Also, aforementioned in \S\ref{sec:training},
we then trained our parsers on the concatenation of each language's treebanks, so that we could parse at the language level (as opposed to treebank). Though we observed scant differences between the two preprocessing pipelines, it was \textsc{udpipe} that produced fewer validation errors. As such, we adopted it as the main preprocessor for our official submission. 

It is apparent in Table~\ref{table:main_results} that the unconnected graph issue (described in \S\ref{sec:postprocessing}) severely affected our official submission to the shared task (observed in the \textit{official} column). After diagnosing and fixing this problem, we observed an improvement of 14.1 ELAS, which is consistent with our scores on the treebanks' development sets. With this in mind, our (fixed) parser tends to perform in a generally stable fashion across languages, with an average ELAS of 76.48 and standard deviation of 6.86. Among our highest scoring languages are Bulgarian, French, and Italian---the former two of which are corroborated by the strong \textit{B1} baseline. Indeed, Tamil is the notable exception among all results, with 56.85 ELAS. We surmise that treebank size is the biggest factor in this degradation of performance, as Tamil has, by far, the smallest treebank at 400 sentences. As such, our parser has comparatively fewer graph samples to train on than it would for some other languages. 

When comparing against the organizer-provided baselines, we see a strong improvement in using our system over both \textit{B2} and \textit{B3} systems. This is encouraging, as it demonstrates the benefit of parsing enhanced dependency graphs directly, rather than relying on predicted trees to accurately relay the enhanced structure (\textit{B1}) or employing a heuristic-driven post-processor to derive it (\textit{B2}). Furthermore, though the organizers consider \textit{B1} as an indirect upper bound due to the gold-standard tokenization and dependency structure employed therein, we can nonetheless observe an advantage in using our parser for some languages. These are Arabic (+2.16 ELAS), Finnish (+3.32), Italian (+4.46), and Ukranian (+0.86). Again, this is promising, given that our parser does not rely on any tree structure in order to parse graphs. 

\subsection{Pre-processing Analysis}

Since the test data was provided in a raw, untokenized format, we were interested in the extent of accuracy loss we might observe when relying on off-the-shelf pre-processors. Table \ref{table:dev_results} displays these results over the development data. It is clear that when we employ predicted segmentation, etc. using either \textsc{stanza} or \textsc{udpipe} pipelines, we observe a slight degradation in accuracy, as compared to the gold data. Omitting Czech, Estonian, Dutch, and Polish (which had several associated treebanks), all other languages degrade by an average of 2.00 ELAS for \textsc{stanza} and 2.32 for \textsc{udpipe}. Though one typically expects such a degradation when evaluating with predicted segmentation, we did observe some unreasonably large gaps in accuracy: namely for Arabic ($-4.02$, $-8.32$ for \textsc{stanza} and \textsc{udpipe}, respectively) and Tamil ($-12.19$, $-8.59$). The latter can likely be explained via its small training set, which undoubtedly affects all components of the preprocessing pipeline. 

When we examine the scores for all multi-treebank languages, we do not notice a large difference between gold and predicted tokenization---which we expect to be different across treebanks. Here, we simply choose the one trained on the largest treebank (\texttt{FicTree} for Czech, \texttt{EDT} for Estonian, \texttt{Alpino} for Dutch, and \texttt{LFG} for Polish), as we consider this a simple yet reliable heuristic. However, when generating predictions for the smaller treebanks using the bigger treebank's preprocessing model, we only notice a notable drop in accuracy for Dutch ($-2.15$, $-2.54$ for \textsc{stanza} and \textsc{udpipe}, respectively). This indicates that there are likely major differences in the treebanks' domains or how they are respectively segmented or annotated. In general, however, the differences between gold and predicted tokenization is surprisingly not as large as we expected. 

\begin{table}[t]
\centering
\begin{tabular}{@{}rccc@{}}
\toprule
          \textbf{Language} & \textsc{stanza} & \textsc{udpipe} & \textbf{Gold Tok.} \\ \midrule
\small\textsc{Arabic}     & 73.66        & 69.36        & 77.68          \\
\small\textsc{Bulgarian}  & 83.46        & 83.17        & 83.89          \\
\small\textsc{Czech}      & 75.60        & 75.47        & 76.00          \\
\small\textsc{Dutch}      & 78.66        & 78.27        & 80.81          \\
\small\textsc{English}    & 80.79        & 79.80        & 82.77          \\
\small\textsc{Estonian}   & 75.43        & 75.32        & 75.81          \\
\small\textsc{Finnish}    & 80.87        & 80.59        & 81.89          \\
\small\textsc{French}     & 86.05        & 85.29        & 88.97          \\
\small\textsc{Italian}    & 85.24        & 85.04        & 85.52          \\
\small\textsc{Latvian}    & 79.00        & 78.39        & 79.28          \\
\small\textsc{Lithuanian} & 74.92        & 74.84        & 75.51          \\
\small\textsc{Polish}     & 71.94        & 73.22        & 73.63          \\
\small\textsc{Russian}    & 78.53        & 78.60        & 78.87          \\
\small\textsc{Slovak}     & 77.54        & 77.33        & 79.17          \\
\small\textsc{Swedish}    & 78.26        & 78.18        & 78.37          \\
\small\textsc{Tamil}      & 50.66        & 54.26        & 62.85          \\
\small\textsc{Ukrainian}  & 79.70        & 79.67        & 79.89          \\ \midrule
\small\textsc{\textbf{Average}}    & 77.08        & 76.87        & 78.88          \\ \bottomrule
\end{tabular}

\caption{Development ELAS for our \textit{fixed} parser.
While in all cases we train the parser on the concatenation of all of a language's gold treebanks (applicable only to Czech, Dutch, Estonian, and Polish),
\textsc{stanza} and \textsc{udpipe} refer to generating predictions on the development data preprocessed by the corresponding pipeline. \textbf{Gold Tok.} refers to generating predictions over gold development data (tokenization, etc).}
\label{table:dev_results}
\end{table}

\section{Conclusion}
\label{sec:conclusion}

In this paper, we have described the IWPT 2020 Shared Task
submission by the Copenhagen-Uppsala team,
consisting of graphs predicted by a transition-based neural
dependency graph parser with pre-trained multilingual contextualized embeddings. While not ranked among the top submission according to the official
scores, the parser architecture proved effective for the type of
dependency graphs exhibited by ED,
and after fixing a critial bug we found the scores
to improve dramatically and agree with the scores we had
observed during development.

We expect that with more resources for BERT fine-tuning,
hyperparameter tuning,
language-specific pre-trained representations and careful
pre- and post-processing, our parser will be a competitive system
in this task.
However, our contribution is a transition system that can directly handle ED, in a unified transition-based parsing system.

 \section*{Acknowledgments}
 We thank the anonymous reviewers for their helpful comments. ML is funded by a Google Focused Research Award. We acknowledge the computational resources provided by CSC in Helsinki and Sigma2 in Oslo through NeIC-NLPL (www.nlpl.eu).

\bibliography{anthology,ku-uu-iwpt20}
\bibliographystyle{acl_natbib}

\appendix

\end{document}